\documentclass[journal=jacsat,manuscript=article]{achemso}
\usepackage[version=3]{mhchem} 
\usepackage{multirow}
\usepackage{multicol}

\usepackage{color}
\usepackage{amsmath,amssymb}

\usepackage{kotex}

\author{Soojung Yang}
\affiliation[KAIST]{Department of Chemistry, KAIST, 291 Daehak-ro, Yuseong-gu, Daejeon 34141, Republic of Korea}

\author{Kyung Hoon Lee}
\affiliation[KAIST]{Department of Chemistry, KAIST, 291 Daehak-ro, Yuseong-gu, Daejeon 34141, Republic of Korea}

\author{Seongok Ryu}
\affiliation[AITRICS]{AITRICS, Hyoryoung-ro 77-gil, Seocho-gu, Seoul, Republic of Korea}
\email{seongokryu@aitrics.com}

\title[An \textsf{achemso} demo]
  {A comprehensive study on the prediction reliability of graph neural networks for virtual screening}


\begin{document}

\begin{tocentry}

Some journals require a graphical entry for the Table of Contents.
This should be laid out ``print ready'' so that the sizing of the
text is correct.

Inside the \texttt{tocentry} environment, the font used is Helvetica
8\,pt, as required by \emph{Journal of the American Chemical
Society}.

The surrounding frame is 9\,cm by 3.5\,cm, which is the maximum
permitted for  \emph{Journal of the American Chemical Society}
graphical table of content entries. The box will not resize if the
content is too big: instead it will overflow the edge of the box.

This box and the associated title will always be printed on a
separate page at the end of the document.

\end{tocentry}

\begin{abstract}

Prediction models based on deep neural networks are increasingly gaining attention for fast and accurate virtual screening systems.
For decision makings in virtual screening, researchers find it useful to interpret an output of classification system as probability, since such interpretation allows them to filter out more desirable compounds. 
However, probabilistic interpretation cannot be correct for models that hold over-parameterization problems or inappropriate regularizations, leading to unreliable prediction and decision making.
In this regard, we concern the reliability of neural prediction models on molecular properties, 
especially when models are trained with sparse data points and imbalanced distributions. 
This work aims to propose guidelines for training reliable models, we thus provide methodological details and ablation studies on the following train principles.
We investigate the effects of model architectures, regularization methods, and loss functions on the prediction performance and reliability of classification results. Moreover, we evaluate prediction reliability of models on virtual screening scenario. 
Our result highlights that correct choice of regularization and inference methods is evidently important to achieve high success rate, especially in data imbalanced situation.
All experiments were performed under a single unified model implementation to alleviate external randomness in model training and to enable precise comparison of results.
\end{abstract}

\section{Introduction}
Recent advancements in deep learning\cite{lecun2015deep} have opened the door to enjoying a variety of unmet molecular applications.
Deep neural networks enables effective task solving thanks to well-designed model architectures suitable for dealing with structural inputs\cite{duvenaud2015convolutional, kearnes2016molecular, gilmer2017neural, wu2018moleculenet}. 
In contrast to using structural or physicochemical descriptors, such as Morgan fingerprints\cite{rogers2010extended}, neural models employ unleashed structural inputs (e.g simplified input molecular line-entry system; SMILES and molecular graph), map them to hidden representations, and make predictions.
For the purpose, convolutional/recurrent neural networks\cite{krizhevsky2012imagenet, kim2014convolutional, hochreiter1997long, cho2014learning} and graph neural networks\cite{scarselli2008graph, battaglia2018relational} have been applied for processing SMILES and molecular graph inputs, respectively.
To this end, they reach to wide range of chemistry problems such as property predictions\cite{duvenaud2015convolutional, kearnes2016molecular, gilmer2017neural, wu2018moleculenet, ryu2019bayesian, zhang2019bayesian}, \textit{de novo} molecular generations\cite{segler2017generating, gomez2018automatic, de2018molgan, sanchez2018inverse, zhavoronkov2019deep, hong2019molecular}, and chemical synthesis planning\cite{segler2018planning, coley2019graph, schwaller2019molecular, dai2019retrosynthesis}.

Albeit with great success, there are key challenges in developing accurate and reliable prediction models arisen from the nature of statistical learning.
Since modern neural networks consist of a large number of parameters, the performance of neural models significantly deteriorates unless a large amount of data is secured\cite{vapnik2013nature, zhang2016understanding}. 
Furthermore, they are prone to make over-confident predictions, in that predictive output is higher than true accuracy.\cite{guo2017calibration}
For example of binary classification problems, the final output of neural networks is produced by a sigmoid activation and is bounded from zero to one.
Hence ones tend to interpret the final output as a probability of belonging to a target class.
If an output of a perfectly calibrated model is 0.8, then ones will interpret that the predictive label is positive with 80\% probability of correctness. 
Such probabilistic interpretation enables ones to rely on the final model output for selecting compounds expected more likely to belong to target class.
However, over-confident model's actual accuracy may be lower than 80\% for predictions with an output probability value of 0.8, and such discrepancy may eventually interrupt the robust decision making.

Therefore, a lot of attempts in vision recognition and language understanding have been made to enhance the reliability as well as performance of model predictions.\cite{snoek2019can, thulasidasan2019mixup} 
For that purpose, regularization methods\cite{srivastava2014dropout}, data augmentations\cite{zhang2017mixup} and advanced learning algorithms\cite{gal2016uncertainty, lakshminarayanan2017simple} have been adopted. 
Previous works\cite{zhang2019bayesian, ryu2019bayesian, schwaller2019molecular} shed light on needs for reliable-AI by studying uncertainty estimation for prediction tasks and chemical reaction planning. 
However, to the best of our knowledge, there is no research with thorough ablation studies that comprehensively study the affect of various factors -- model architectures, regularizations, and learning and inference algorithms -- on prediction reliability. 
This motivates us to start this work.

In particular, models that speak their results in the language of probability allows us to choose more desirable compounds in virtual screening, which will then be taken into account for experimental validation. 
One common approach is to select samples with high predictive output (sometimes referred to as confidence).
\citeauthor{stokes2020deep}\cite{stokes2020deep} screened compounds from drug repurposing hub library by using the prediction score of the ensemble model and experimentally validated their efficacy. 
In order to enhance success rate of virtual screening, however, ones may need to evaluate whether the model gives true probability of correct prediction, i.e. the relationship between true accuracy and prediction confidence. 
We point out that current evaluations are limited to validating model performance and providing averaged scores on entire data points of test sets, however, does not evaluate the model in terms of prediction reliability. 

In this work, we present a comprehensive study on the reliability of prediction models based on graph neural network\cite{battaglia2018relational, kipf2016semi, velivckovic2017graph, duvenaud2015convolutional} in classification tasks. 
We focus on how to assess and improve prediction reliability in order for successful virtual screening with probabilstic interpretation of final outputs to be possible. 
The rest of paper firstly provides preliminaries on strategies to evaluate prediction reliability, i.e. calibration curve, expected calibration error, and entropy histogram. 
Then, we briefly introduce methods in our scope -- graph convolutional network and its augmentations, regularizations, and focal loss.
Numerical experiments investigate the affect of model architectures, regularizations, and also their implications on virtual screening. 
Our study leaves lessons that relevant model capacity and appropriate regularizations is key to achieve high success rate in screening more desirable compounds with prediction probability.

\section{Preliminaries on prediction reliability}

We elaborate the methods to evaluate prediction reliability.
Let us write our model $f$ produces an output $\hat{p}_i = f(\textbf{x}_i)$ for a given input $\textbf{x}_i$. 
Then, a predictive label $\hat{y}_i$ is determined by the threshold-based estimator:
\begin{equation}
    \hat{y_i} = 
    \begin{cases}
        1 & \text{if } \hat{p}_i > \delta \\
        0 & \text{otherwise},
    \end{cases}
\end{equation}
where $\delta$ is the threshold, and 0.5 is usually chosen. 
If ones would like to interpret the final output $\hat{p_i}$ as a true confidence (or probability) of correct prediction $p_i$, 
a model should be \textbf{perfectly calibrated}.
As proposed in \citeauthor{guo2017calibration}, a \textbf{perfect calibration} of models can be defined as follows:
\begin{equation}
    P(\hat{Y}=y | \hat{P}=p) = p, \quad \forall p \in [0,1].
\end{equation}
They also defined the term \textbf{expected calibration error (ECE)}, 
\begin{equation}
    \text{ECE} = \mathbb{E}_{\hat{P}}[\lvert P(\hat{Y}=y | \hat{P}=p) - p \rvert],
\end{equation}
which can be interpreted as the gap between true and model's confidence. 
We will introduce the empirical ECE estimator later.

In order to evaluate the reliability (calibration performance) of models with empirical data points, 
we utilize \textbf{calibration curve}, \textbf{expected calibration error (ECE)}, and \textbf{entropy histogram}. 
If we divide the predictive results into the total $M$ number of bins (intervals), 
then the accuracy and the confidence of predictions in the $m$-th bin $B_m$ is given by 
\begin{equation}
    \text{acc}(B_m) = \frac{1}{|B_m|} \sum_{i \in B_m} \mathbb{I}(\hat{y}_i = y_i),
\end{equation}
and
\begin{equation}
    \text{conf}(B_m) = \frac{1}{|B_m|} \sum_{i \in B_m} \hat{p}_i,
\end{equation}
where $|B_m|$ is the number of samples in $B_m$, and $\mathbb{I}$ is an indicator function. 
\textbf{Calibration curve} visualizes $\text{conf}(B_m)$ and $\text{acc}(B_m)$ for all bins $m \in [0, ..., M]$, as shown in Figure \ref{fig:exp1_summary}, \ref{fig:exp2_summary} and \ref{fig:exp4_plots}. 
Ones can estimate the calibration error of each bin by computing the gap between the perfect calibration curve and the accuracy-confidence curve.
So as to, \textbf{ECE} summarizes the calibration errors over entire data points, whose estimator is given by 
\begin{equation}
    \text{ECE} = \sum_{m=1}^{M} \frac{|B_m|}{n} |\text{acc}(B_m) - \text{conf}(B_m)|.
\end{equation}

Lastly, we also provide the distribution (histogram) of \textbf{predictive entropy}, which is defined as
\begin{equation}
    H(p) = -p \log p - (1-p) \log (1-p), \quad   \forall p \in (0, 1),
\end{equation}
Note that predictive entropy represents the amount of information lacks in predictions, in other words, predictive uncertainty. 
That being said, if a model does not have enough information on samples, predictions will show high predictive entropy.
But, over-confident models tend to show large amount of zero entropy predictions and vice versa, as shown in our experimental demonstration. 
We note that predictive entropy is maximum at $p=0.5$ and minimum at $p = 0.0$ or $1.0$. 

\section{Methods}
In this section, we describe the methods -- model architectures, regularization methods, and loss functions -- whose effects on prediction performance and reliability are investigated with numerical experiments. 

\subsection{Model architectures}

We express molecular graphs with undirected graph $G(X,A)$, where $X \in \mathbb{R}^{N \times d}$ is a set of $N$ node features, and $A \in \mathbb{R}^{N \times N}$ is an adjacency matrix. Note that we consider connectivity between nodes only, i.e. $A_{ij} \in {0,1}$ for all node pairs $(i,j)$. 
Graph neural networks (GNNs) for graph-level prediction tasks consist of three parts: i) an encoder featurizes input node information, ii) a readout summarizes node features and produces graph features, and iii) a predictor maps graph features to target property values. 
Among the various GNN variants, we consider a \textbf{graph convolutional network (GCN)}\cite{kipf2016semi, duvenaud2015convolutional} as a baseline 
and its advancements augmented with self-attention mechanism in node and/or graph featurizations. 

A simple expression on node featurizations in GCN is given by 
\begin{equation}
    \tilde{H}^{l+1} = \text{ReLU}(AH^{l}W^{l}),
\end{equation}
where $H^{l} \in \mathbb{R}^{N \times d^{l}}$ is a set of node features, which have $d^{l}$ dimension for the $l$-th graph convolution layer, $H^{0} = X$, $W^{l} \in \mathbb{R}^{d^{l} \times d^{l+1}}$ is a weight parameter, and $\text{ReLU}$ is rectifier linear unit (ReLU) activation.
Graph convolution layer can be improved by applying attention mechanism\cite{vaswani2017attention} that computes attention coefficients between a set of query and key node feature pairs. 
By following the analogy in \textbf{graph attention network (GAT)} \cite{velivckovic2017graph}, graph attention layer updates node features by
\begin{equation}
\label{eqn:gat}
    \tilde{H}_i^{l+1} = \text{ReLU}(\sum_{j \in \mathcal{N}_i} \alpha_{ij}^{l} H_j^{l}W^{l}),
\end{equation}
where $\mathcal{N}_i$ denotes a set of adjacent nodes and $i$-th node itself, $H_{i}^{l}$ denotes $i$-th node feature
and $\alpha_{ij}^{l} = f(H_i^{l}W^{l}, H_j^{l}W^{l}) \in \mathbb{R}$ is attention coefficient whose query and key vectors are $H_i^{l}W^{l}$ and $H_j^{l}W^{l}$ respectively. 
We adopt the self-attention mechanism\cite{vaswani2017attention} to compute the attention coefficient between adjacent node features:
\begin{equation}
\label{eqn:gat_coeff}
    \alpha_{ij} = \text{tanh}(\frac{(H_i^{l}W^{l}) W_a^{l} (H_j^{l}W^{l})^T}{\sqrt{d^{l+1}}}),
\end{equation}
where $W_{a}^{l} \in \mathbb{R}^{d^l \times d^l}$ is a weight parameter and $\text{tanh}$ is a hyperbolic-tangent activation.
Note that dividing the dot-product output by the scaling factor $\sqrt{d^{l+1}}$ significantly stabilizes training via stochastic gradient descents as explained in \citeauthor{vaswani2017attention}\cite{vaswani2017attention}.
While GAT\cite{velivckovic2017graph} used softmax activation for the nonlinearity $\tau$, 
we empirically found that tanh activation works better than softmax activation for our tasks. 
(We breifly discuss the matter of choosing proper activation function in supplementary information.) 
Based on the above node featurizations, we compose each node embedding block with a graph convolution/attention layer, a dropout layer and a residual connection\cite{he2016deep}, i.e. $H^{l+1} = \tilde{H}^{l+1} + H^{l}$.

A readout layer aggregates a set of node features and returns a graph feature vector $z^{l} \in \mathbb{R}^{d_g^l}$. 
We added subscript $g$ (graph) to $z, W$ and $d$ for weight parameters in readout layers, to emphasize that they are different set of weight parameters to convolution layers.
The most basic operation that satisfy permutation invariance for aggregation is \textbf{summation-readout}
\begin{equation}
    z_g^{l} = \text{sigmoid}(\sum_{i=1}^{N} H_i^{l}W_g^{l}),
\end{equation}
where $W_g^{l} \in \mathbb{R}^{d^{l} \times d_g^l}$ is a weight parameter.
Beyond summarizing node features with equal weights, it would be more powerful to aggregate node features with different importances.
For this purpose, we adopt self-attention in the readout step again, as proposed in \citeauthor{lee2019set}\cite{lee2019set}:
\begin{equation}
    z_g^{l} = \text{sigmoid}(\sum_{i=1}^{N} \alpha_{i}^{l} H_i^{l}W_g^{l}),
\end{equation}
where the attention coefficient $\alpha_{i}^{l}$ is given by
\begin{equation}
\label{eqn:attention_readout}
    \alpha_i^{l} = N \times \text{softmax}(\frac{\textbf{1} (H_i^{l}W_g^{l})^T}{\sqrt{d_g^{l}}}),
\end{equation}
where $\textbf{1} \in \mathbb{R}^{d_g^{l}}$ is a vector whose elements are one. 
This \textbf{attention readout} computes similarity between the one-vector (query vector) and the $l$-th node features (key vectors), and uses resulting coefficient for linear-combination of node features (value vectors).
To aggregate with appropriate summary statistics, we scale the attention coefficient with the number of node features $N$ after applying softmax activation. 
We experimentally found that this scaling allows the outputs of the attention readout to be distinguishable for given two different graphs. 
We discuss this fact in supplementary information.
We use the concatenation of all the outputs of the $l$-th readout layers for $l \in [1, ..., L]$, as proposed in \citeauthor{xu2018powerful}\cite{xu2018powerful}:
\begin{equation}
    z_{G} = \text{CONCAT}([z^1, ..., z^{L}]).
\end{equation}
where $L$ is the number of node embedding layers.
Since the outputs of $l$-th graph convolution layer can be thought as the $l$-hop substructure of center nodes, this concatenation enables the predictor to use the hierarchical structures of input graphs.
A linear classifier computes the final output by using a graph feature input
\begin{equation}
    \hat{p} = \text{sigmoid}(z_{g}^{T} W_{c}  + b_{c}),
\end{equation}
where $W_{c}$ and $b_{c}$ are weight and bias parameters for the classifier.

\subsection{Regularizations}

Regularizing neural networks is obviously important to prevent over-fitting problem, which degrades prediction performance. 
Furthermore, they can lead to obtain well-calibrated high prediction probability.  
In this section, we introduce regularization methods widely used in modern neural networks and our experimental investigation as well.

\textbf{Dropout}\cite{srivastava2014dropout} is one of the most popular regularization methods. 
Its first proposal interpreted the effect of dropout as preventing models to be dependent on specific input or hidden features. 
Furthermore, \citeauthor{gal2016dropout}\cite{gal2016dropout} proposed \textbf{Monte Carlo-dropout (MC-DO)}, approximate Bayesian inference method with dropout variational posterior. Its predictive distribution is given by the MC sampling of outputs produced by model parameters with stochastic dropout masks. 
In our experiments, we both investigate the effect of \textbf{standard dropout (DO)}, which does not use stochastic dropout mask in test phase, and \textbf{MC-dropout} .  

\textbf{Label smoothing (LS)}\cite{szegedy2017inception} is a simple regularization method that add a small uniform perturbations to each $c$-th class label $y_{i,c}$ for the input $\textbf{x}_i$.
The perturbed labels of the $c$-th class $y_{i,c}^{\text{LS}}$ for the input $\textbf{x}_i$ is given by 
\begin{equation}
\label{eqn:LS}
	y_{i,c}^{LS} = y_{i,c} (1-\alpha_{\text{LS}}) + \frac{\alpha_{\text{LS}}}{C}, \quad \forall c \in {1,...,C}
\end{equation}
where $\alpha_{\text{LS}}$ is the amount of perturbation, and $C$ is the number of classes. 
Note that all the experiments in this study are binary classification, i.e. $C=2$. 

The learning objective of training with LS is given by 
\begin{equation}
\label{eqn:objective_LS}
    \mathcal{L}_{\text{LS}}(\textbf{y}, \hat{\textbf{p}};\alpha_{\text{LS}}) = \mathcal{L}_{\text{BCE}}(\textbf{y}^{\text{LS}}, \hat{\textbf{p}}).
\end{equation}
where $\mathcal{L}_{\text{BCE}}(\textbf{y}, \hat{\textbf{p}}) =  \sum_{i=1}^{n} -y_i  \log \hat{p}_i - (1-y_i) \log (1-\hat{p}_i)$ is binary cross-entropy (BCE) loss.

\textbf{Entropy regularization (ERL)}\cite{pereyra2017regularizing} is a regularization method to penalize over-confident predictions, by introducing the predictive entropy $H(\hat{p})$ as a penalty term, like the way in L2-weight decay.
The learning objective with ERL is given by 
\begin{equation}
\label{eqn:ERL}
    \mathcal{L}_{\text{ERL}}(\textbf{y}, \hat{\textbf{p}}; \beta) = \mathcal{L}_{\text{BCE}}(\textbf{y}, \hat{\textbf{p}}) - \beta \sum_{i=1}^{n}  H(\hat{p}_i),
\end{equation}
where $\beta$ is a hyper-parameter that controls the amount of predictive entropy penalty.

\subsection{Focal loss}

\textbf{Focal loss (FL)}\cite{lin2017focal} is a well-known loss function for detecting rare samples in imbalanced data distribution by penalizing predictions of high output probability . 
While the learning objective with ERL is given by the summation of original loss function (i.e. BCE) and its regularization term, 
the learning objective with the FL is simply given by: 
\begin{equation}
\label{eqn:FL}
    \mathcal{L}_{\text{FL}}(\textbf{y}, \hat{\textbf{p}}; \gamma_{\text{FL}}) = \sum_{i=1}^{n}- y_i(1 - \hat{p}_i)^{\gamma_{\text{FL}}} \log \hat{p}_i - (1-y_i) \hat{p}_i^{\gamma_{\text{FL}}} \log (1-\hat{p}_i),
\end{equation}
without any additional penalty, where weights depend on the output of the neural network. $\gamma_{\text{FL}} > 0$ is a hyperparameter that controls an extent that the over-confidence is penalized. 
The factor $(1-p_{i})^{\gamma_{\text{FL}}}$ in the first term of R.H.S. reduces $\log \hat{p}_{i}$ significantly for large value of $p_{i}$ (near to 1). 
On the other hand, $\hat{p}_{i}^{\gamma_{\text{FL}}}$ in the 2nd term of R.H.S reduces $\log (1-p_{i})$ significantly for small value of $p_{i}$ (near to 0). 
As a result, training with the focal loss penalizes the over-confident predictions by enforcing the output to be less confident (output $p_{i}$ are not near to either 0 or 1).

As proposed in \citeauthor{lin2017focal}\cite{lin2017focal}, we performed the experiments with weighted focal loss (WFL), given by
\begin{equation}
\label{eqn:FL}
    \mathcal{L}_{\text{WFL}}(\textbf{y}, \hat{\textbf{p}}; \alpha_{\text{FL}}, \gamma_{\text{FL}})
    = \sum_{i=1}^{n} - \alpha_{\text{FL}} y_i (1-\hat{p}_i)^{\gamma_{\text{FL}}} \log \hat{p}_i - (1-\alpha_{\text{FL}}) (1-y_i)\hat{p}_i^{\gamma_{\text{FL}}} \log (1-\hat{p}_i),
\end{equation}
where $\alpha_{\text{FL}}$ and $(1-\alpha_{\text{FL}})$ are hyper-parameters that role as weight factors for the prediction losses on positive and negative samples, respectively. 

\subsection{Interpretation of the effect of cost-sensitive learning}
In this section, we describe how cost-sensitive learning (i.e. LS, ERL, and FL) is interpreted as a regularization of probability estimator.
To this end, we conclude that cost-sensitive learning could not provide well-calibrated results, but biased probability estimation

The learning objective of training with LS can be rewritten as 
\begin{equation}
\label{eqn:LS_KLD}
    \mathcal{L}_{LS}(\textbf{y}, \hat{\textbf{p}}; \alpha_{\text{LS}}) = \mathcal{L}_{\text{BCE}}(\textbf{y}, \hat{\textbf{p}}) + \beta \sum_{i=1}^{n} \text{KL}[\mathcal{U}(y_i) \Vert \hat{P}(y_i|\textbf{x}_i)] + \text{const.},
\end{equation}
where $\text{KL}[p \Vert q]$ is the Kullback-Leibler (KL) divergence between two distributions $p$ and $q$, and $\mathcal{U}(y)$ denotes the uniform distribution. 
It is straightforward to show eqn. \ref{eqn:LS_KLD} becomes equivalent to eqn. \ref{eqn:objective_LS}.
By using the definition of KL divergence, the penalty term (the second term of R.H.S.) in eqn. \ref{eqn:LS_KLD} is given by
\begin{equation}
\label{eqn:LS_penalty}
\begin{split}
    \text{KL}[\mathcal{U}(y_i) \Vert \hat{P}(y_i|\textbf{x}_i)] 
    &= \mathbb{E}_{\mathcal{U}(y_i)} [ \log \frac{\mathcal{U}(y_i)}{\hat{P}(y_i|\textbf{x}_i)}] \\
    &= - \sum_{c=1}^{C} \frac{1}{C} \log \hat{p}_{i,c} + \frac{1}{C} \log \frac{1}{C}.
\end{split}
\end{equation}
If we let $\beta$ in \ref{eqn:LS_KLD} as the constant multiple of $\alpha_{\text{LS}}$, it concludes the proof.

Similarly, the learning objective of training with ERL can be rewritten as 
\begin{equation}
\label{eqn:ERL_KLD}
        \mathcal{L}_{\text{ERL}}(\textbf{y}, \hat{\textbf{p}}; \beta) 
        = \mathcal{L}_{\text{BCE}}(\textbf{y}, \hat{\textbf{p}}) + \beta \sum_{i=1}^{n} \text{KL}[P_\theta(y|\textbf{x}) \Vert \mathcal{U}(y)] + \text{const.},
\end{equation}
and the penalty term also can be rewritten as 
\begin{equation}
\label{eqn:ERL_penalty}
\begin{split}
    \text{KL}[\hat{P}(y_i|\textbf{x}_i) \Vert \mathcal{U}(y_i)] 
    &= \mathbb{E}_{\hat{P}(y_i|\textbf{x}_i)} [ \log \frac{\hat{P}(y_i|\textbf{x}_i)}{\mathcal{U}(y_i)}] \\
    &= +\sum_{c=1}^{C} \hat{p}_{i,c} \log \hat{p}_{i,c} - \hat{p}_{i,c} \log \frac{1}{C}.  
\end{split}
\end{equation}
Since $\sum_{i=1}^{n} \sum_{c=1}^{C} \hat{p}_{i,c} \log C = n \log C$ is constant, we confirm that eqn. \ref{eqn:ERL} is equivalent to eqn. \ref{eqn:ERL_KLD}. 

We can understand that LS and ERL penalize confident predictions by enforcing the prediction distribution to the uniform distribution. 
The key difference between LS and ERL is that the former and the latter minimize the forward and reverse KL-divergences, respectively. 
As a result, LS penalizes all predictions with equal weight (i.e. $1/C$), on the other hand, ERL penalizes over-confident predictions with larger weight (i.e. $\hat{p}_{i,c}$). 

Lastly, we interpret the learning objective of FL as the BCE with asymmetric entropy regularization.
For our understanding, we use the approximate relation $(1-\hat{p})^{\gamma_{\text{FL}}} \approx 1 - \gamma_{\text{FL}} \hat{p}$, and we can rewrite the FL as follows:
\begin{equation}
\begin{split}
    \mathcal{L}_{\text{FL}}(\textbf{y}, \hat{\textbf{p}}; \gamma_{\text{FL}}) 
    &\approx - \sum_{i=1}^{n} y_i \log \hat{p}_i - (1-y_i) \log (1-\hat{p}_i) \\
    & - \gamma_{\text{FL}} \{-y_i \hat{p}_i \log \hat{p}_i - (1-y_i) (1-\hat{p}_i) \log (1-\hat{p}_i)  \}  \\
    &= \mathcal{L}_{\text{BCE}}(\textbf{y}, \hat{\textbf{p}}) - \gamma_{\text{FL}} \sum_{i=1}^{n} H_{\text{asym}} (y_i, \hat{p}_i),
\end{split}
\end{equation}
where asymmetric entropy is defined as $H_{\text{asym}} (y_i, \hat{p}_i) = -y_i \hat{p}_i \log \hat{p}_i - (1-y_i) (1-\hat{p}_i) \log (1-\hat{p}_i)$.
We can understand that maximizing the asymmetric entropy discourages over-confident prediction on the given true labels, while maximizing the standard entropy (ERL) penalizes regardless of labels. 

The above learning algorithms have their learning objective as a form of $\mathcal{L}_{\text{BCE}}(\textbf{y}, \hat{\textbf{p}}) + \beta f(\hat{\textbf{p}})$, a summation of BCE and predictive probability regularizer. 
Theory of logistic regression reveals that minimizing BCE gives asymptotic convergence of the model output to $p^* = P(Y=1|X)$ -- the probability of observing positive sample given input random variable $X$ -- as a number of training (empirical) samples increases.
On the other hand, cost-sensitive learnings introduce additive probability regularizer and enforce predictive distribution to be similar with uniform distribution. 
It can help to alleviate over-confident prediction since it maximizes predictive entropy,
but does not guarantee the convergence of output to unbiased probability estimation. 
Previous works\cite{szegedy2017inception, muller2019does, pereyra2017regularizing, lin2017focal, thulasidasan2019mixup} in other domains empirically show that cost-sensitive learning can improve predictive performance and/or reliability. 
We show yet undiscovered results in molecular property prediction tasks, emphasizing the importance of appropriate regularizers for well-calibrated probability estimation.

\section{Experiments}

\subsection{Dataset - BACE, BBBP, HIV}

\begin{table}[]
    \begin{tabular}{|c|c|c|c|c|}
    \hline
    \multicolumn{1}{|l|}{} & BACE          & BBBP         & HIV          \\ \hline
    Task type              & \multicolumn{3}{c|}{Binary classification}  \\ \hline
    Number of samples      & 1,513         & 2,050        & 41,127       \\ \hline
    Positives:Negatives    & 822:691       & 483:1567     & 39684:1443   \\ \hline
    Total training epoches & \multicolumn{2}{c|}{200}     & 100          \\ \hline
    Decay steps            & \multicolumn{2}{c|}{80, 160} & 40, 80       \\ \hline
    \end{tabular}
    \caption{Specifications on datasets and model training in this work.}
    \label{tab:dataset_spec}
\end{table}

We used the three datasets -- BACE, BBBP and HIV sets -- which are widely used in machine learning applications of property predictions.
The BACE dataset provides qualitative (binary label) binding results for set of human inhibitors of human beta-secretase 1. 
The BBBP dataset includes binary labels on the blood-brain barrier permeability properties for chemical compounds. 
The HIV dataset gives binary labels on the ability to inhibit HIV replication.
We obtained input and label pairs from the MoleculeNet homepage.\cite{wu2018moleculenet}

\subsection{Training scheme}
Table \ref{tab:dataset_spec} summarizes the number of samples, task type, the total number of training epoches and decay steps. 
Each dataset was split to training set and test set by 80:20 ratio. 
We used AdamW optimizer\cite{loshchilov2018decoupled} (Adam optimizer\cite{kingma2014adam} with decoupled weight decay) for gradient-descent optimization.
Hyper-parameters such as the total number of training epochs and steps to start learning rate decaying are noted in Table \ref{tab:dataset_spec}.
Initial learning rate was set to $10^{-3}$ and further decayed by the factor of 0.1. 
We set the number of node embedding layers as 4, the dimensions of node features $d^{l}$ as 64 and graph features $d_g^l$ as 256.

Since the number of data points are small, the models were evaluated by averaging the results of five-fold experiments; five sets of train-test split were made with five different random seeds. 
We regularized the model by using L2-weight decay with coefficient $10^{-4}\times(1-p_{\text{do}})$ where $p_{\text{do}}$ is the dropout probability. 
For clear visualization, we show calibration curves, entropy histogram, and output probability histograms each of which is obtained by using the first random seed. 

\subsection{Effect of model architectures on prediction performance and reliability}

Firstly, we investigate the effect of model architectures (parameterizations) on both prediction performance and reliability. 
While related researches thrive in computer vision and natural language understanding fields, this important question still has not been answered with well-designed ablation study in molecular applications.
Thus, we aim to answer the following question: 
``Does the recently invented neural model designs - graph attention network and attention readout - show their promising effect on molecular property prediction?''

\begin{figure}[] 
    \includegraphics[width=0.95\textwidth,trim={0cm 0 0cm 0},clip]{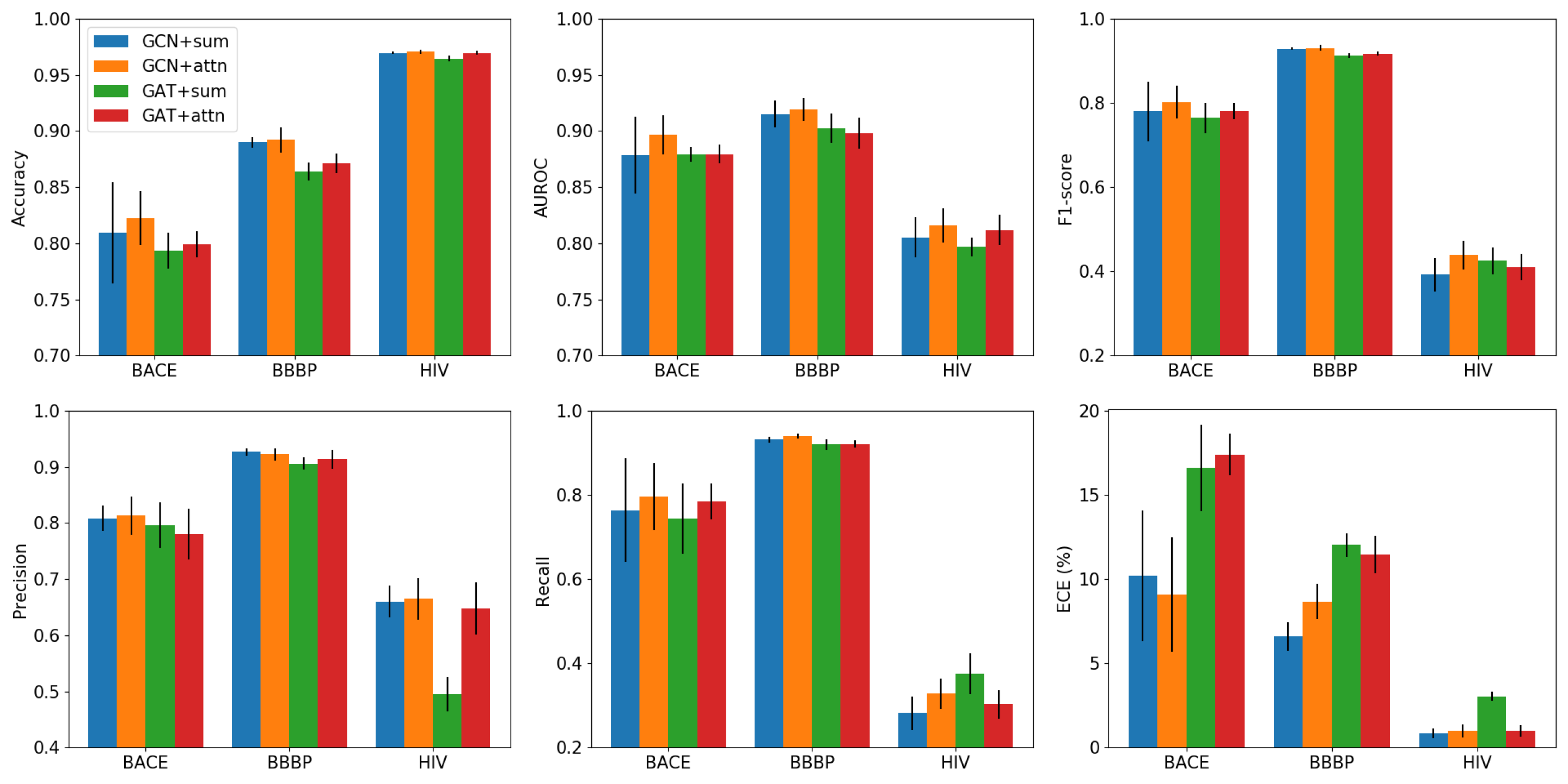}
    \caption{The prediction performance (accuracy, AUROC, precision, recall and F1-score) and reliability (ECE) of four different models. Sum and Attn stand for the summation and attention readouts.}
    \label{fig:exp1_results_summary}
\end{figure}

In Figure \ref{fig:exp1_results_summary}, we summarize the prediction performance and reliability of the four different models -- `GCN with summation readout (GCN+Sum)', `GCN with attention readout (GCN+Attn)', `GAT with summation readout (GAT+sum)', and `GAT with attention readout (GAT+Attn)'. GCN and GAT are used for node featurization before readout.
We found that the 'GCN+Attn' model shows the best prediction performance for the three datasets (tasks). 
Observing the results, GAT seems to degrade both prediction performance and reliability; the usage of GAT significantly harmed ECE, in particular.

\begin{figure}[] 
    \includegraphics[width=0.95\textwidth,trim={0cm 0 0cm 0},clip]{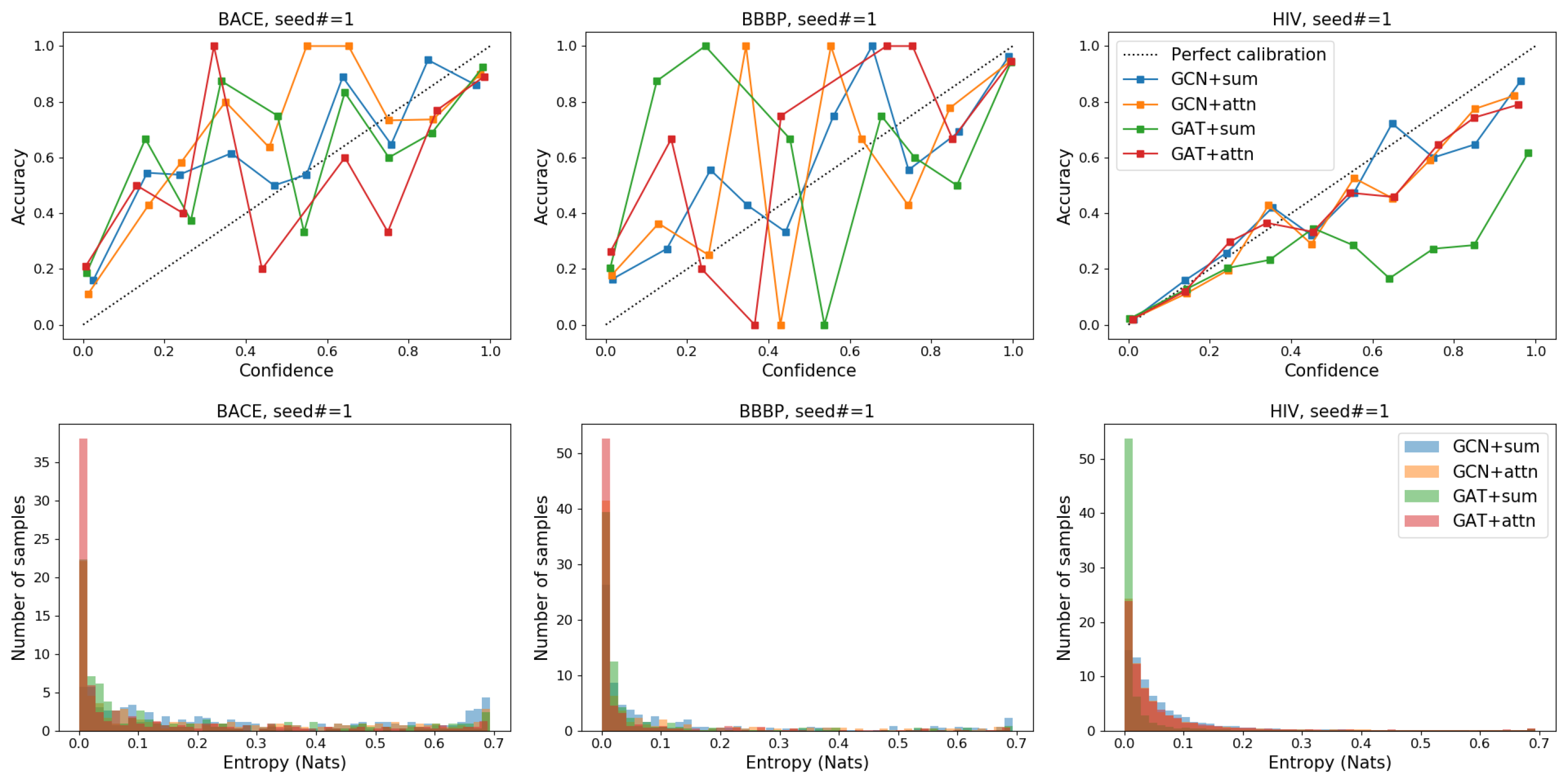}
    \caption{(Top) Calibration curves and (Bottom) predictive entropy histograms of four different models}
    \label{fig:exp1_summary}
\end{figure}

In order to provide interval-wise information of predictions in addition to the results averaged on the entire test set, we visualize the calibration curves and the entropy histograms in \ref{fig:exp1_summary}.
As high ECE values highlight, using GAT significantly enlarged the variance in accuracy ($\text{acc}(B_m)$) across the bins and increased the gap between confidence ($\text{conf}(B_m)$) and accuracy.
Also, the predictive entropy of the GAT models is located near 0.0 more frequently compared to the GCN models. 
With such evidence, we can conclude that GAT models are prone to over-confidence problems. 
This observation tells us that probabilistic interpretation of GAT model's output wouldn't be feasible unless the model is calibrated. Thus, for reliable virtual screening, it is necessary to calibrate the GAT model.
We note that the large variance in predictive accuracy for the BACE and the BBBP tasks might have arisen due to the small size of the datasets. 

We note that the GAT models shows better prediction performance than GCN models for regression tasks with a large amount of data samples - unlike for aforementioned classification tasks - as shown in Figure \ref{fig:num_conv_layers} and Table \ref{tab:zinc_predictions} in supplementary information.
We conjecture that the small number of data samples and imbalanced distribution of classification datasets made GAT models to perform worse, as GAT models consist of more parameters than GCN models.
For the next following ablation studies, testing the effect of regularization methods and focal loss, we set `GCN+Attn' as the baseline model which shows the best performance and reliability results.

\subsection{Effect of regularizations on prediction performance and reliability}

\begin{figure}[] 
    \includegraphics[width=0.95\textwidth,trim={0cm 0 0cm 0},clip]{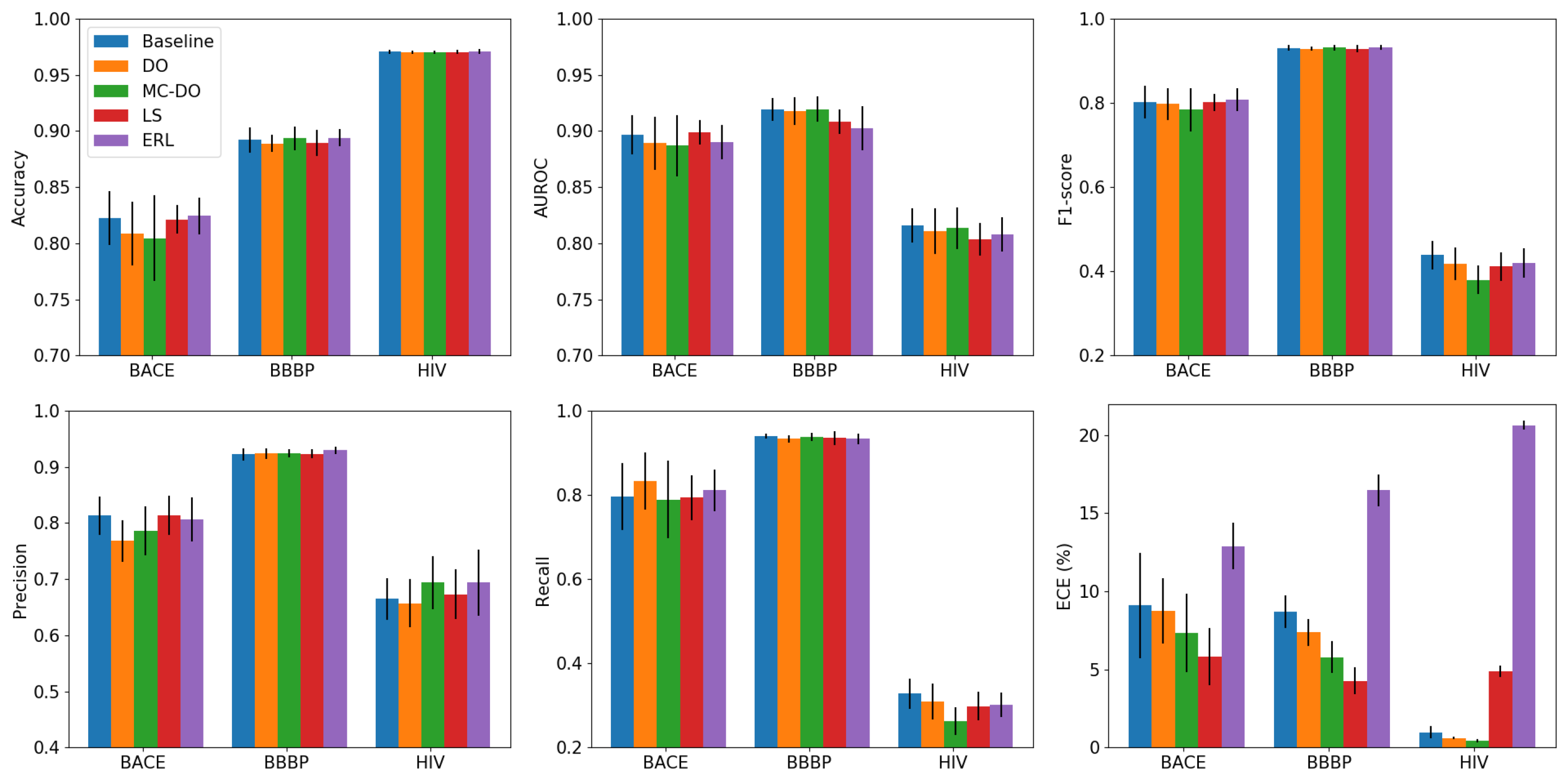}
    \caption{The prediction performance (accuracy, AUROC, precision, recall and F1-score) and reliability (ECE) of the models each of which adopts a different regularization method on the baseline architecture `GCN+Attn'.}
    \label{fig:exp2_results_summary}
\end{figure}

Next, we investigate the effect of well-known regularization methods on enhancing the reliability of our baseline model, i.e. `GCN+Attn'. 
We adopted a number of regularization methods --  standard dropout (DO), Monte-Carlo dropout (MC-DO), label smoothing (LS), and entropy regularization (ERL). 
The hyper-parameters for each method are included in the implementation detail section. 
Figure \ref{fig:exp2_results_summary} summarizes the prediction performance (accuracy, AUROC, and F1-score) and reliability (ECE) of the models implementing above methods
All the models have resulted in similar prediction performance. 
On the other hand, prediction reliability widely varied depending on the regularization methods: applying DO and MC-DO have improved the reliability, while LS and ERL have made it worsen.
We found that MC-DO is more effective than standard DO for all three tasks. 
LS shows the best calibration results for the BACE and BBBP prediction, but it underperforms the baseline model for the HIV prediction. 
ERL underperforms the baseline model for every task. 
According to our theoretical analysis, well-calibrated probability is not granted for LS and ERL. Disappointing performance of LS for HIV prediction and the overall poor performance of ERL can be explained in this regard.

\begin{figure}[] 
    \includegraphics[width=0.95\textwidth,trim={0cm 0 0cm 0},clip]{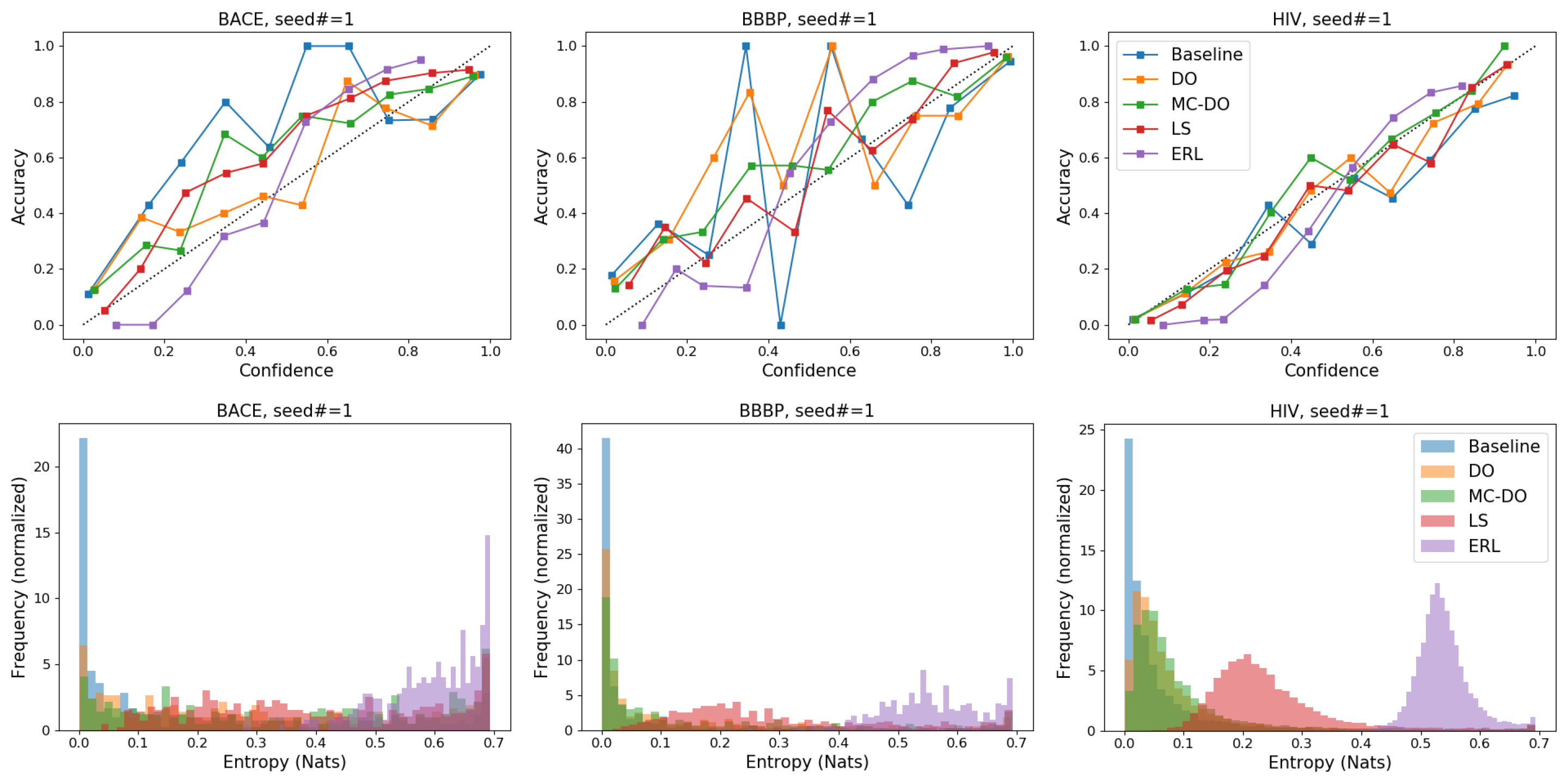}
    \caption{Calibration curves (top) and entropy histograms (bottm) of models using a number of regularization methods for the three different tasks.}
    \label{fig:exp2_summary}
\end{figure}

Figure \ref{fig:exp2_summary} shows the calibration curves and entropy histograms of different regularization methods for BACE, BBBP, and HIV tasks. 
As the lowered ECE values point out, applying DO, MC-DO, and LS has diminished the deviation between the perfect calibration curve (the black dotted line) and the experimental calibration curves.
Such decline in ECE values comes from the suppression of highly over-confident predictions. Notably, MC-DO better regularized the model than standard DO.  
We can obtain more insights by observing the entropy histograms. 
For the baseline model, predictive entropy values are highly frequent at 0.0, implying that the predictive outputs are mostly 0 or 1. 
On the contrary, for LS and ERL models, most of the predictive entropy values are larger than 0.0 - even centered around 0.5 - 0.6 for ERL. 
It seems like ERL is showing an excessive regularization effect, as minimizing the forward KL-divergence between the predictive distribution and an uniform distribution sometimes gives such result.

\subsection{Effect of focal loss on prediction performance and reliability}

A lot of public datasets are imbalanced in that the number of samples from majority and minority class are largely different. 
For example, the ratio of active and inactive compounds in the HIV dataset is 3:97. 
The true distribution might be similarly imbalanced in nature; there are much less bio-active compounds than inactive compounds. 
Focal loss\cite{lin2017focal} has been well-known for treating imbalanced datasets, especially that of image datasets. 
However, to the best of our knowledge, there is no previous work studying the predictive reliability of models adopting focal loss.
Thus, we investigate the effects of focal loss on predictive performance and reliability for the HIV activity detection task.

\begin{figure}[]
\includegraphics[width=0.95\textwidth,trim={0cm 0 0cm 0},clip]{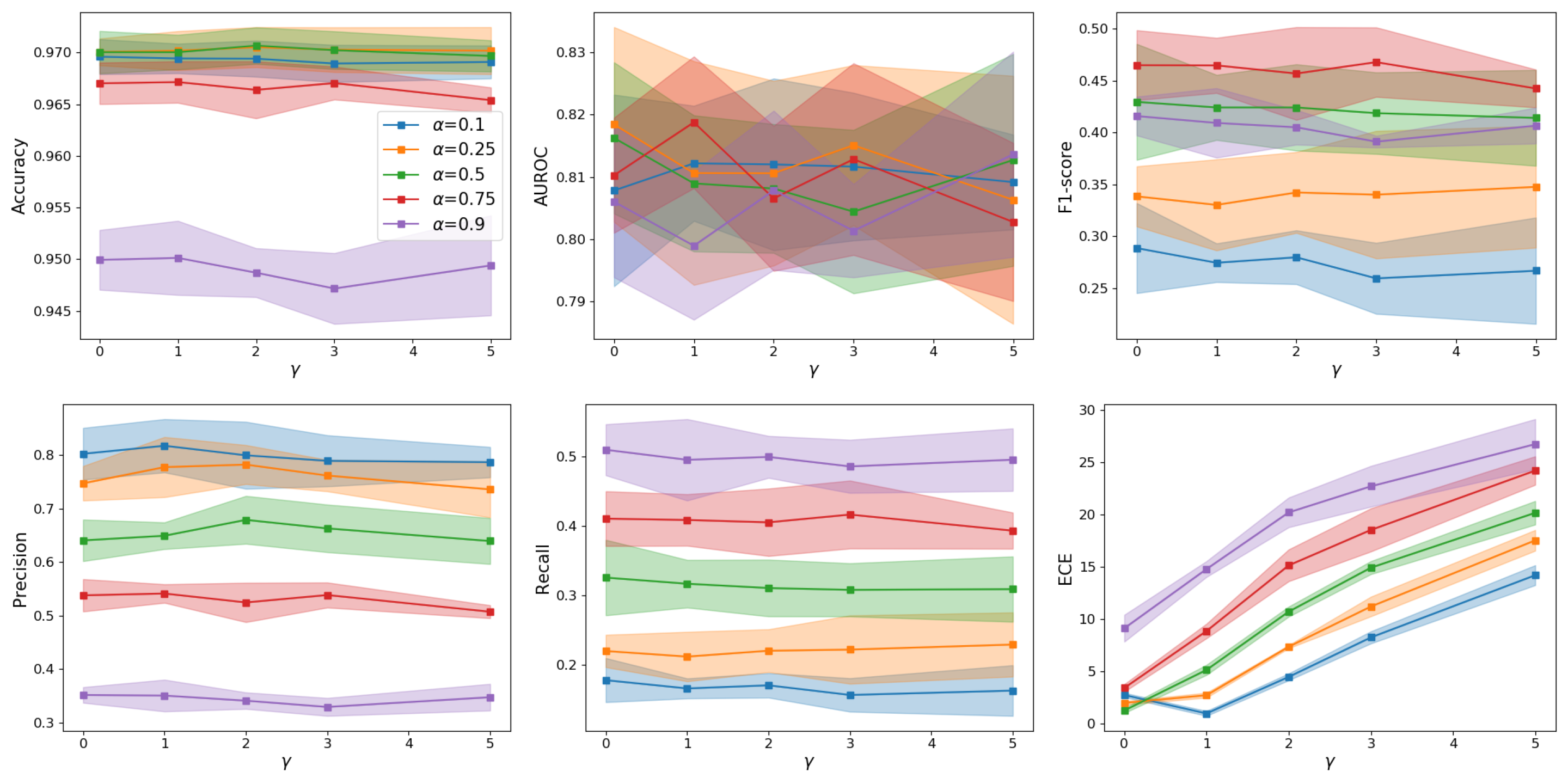}
\caption{The change of the prediction performance (accuracy, precision, recall, and F1-score) and the prediction reliability (ECE) as varying the hyperparameters in the focal loss, i.e. $\alpha_{\text{FL}}$ and $\gamma_{\text{FL}}$.}
\label{fig:exp4_summary}
\end{figure}

\begin{figure}[]
\includegraphics[width=0.95\textwidth,trim={0cm 0 0cm 0},clip]{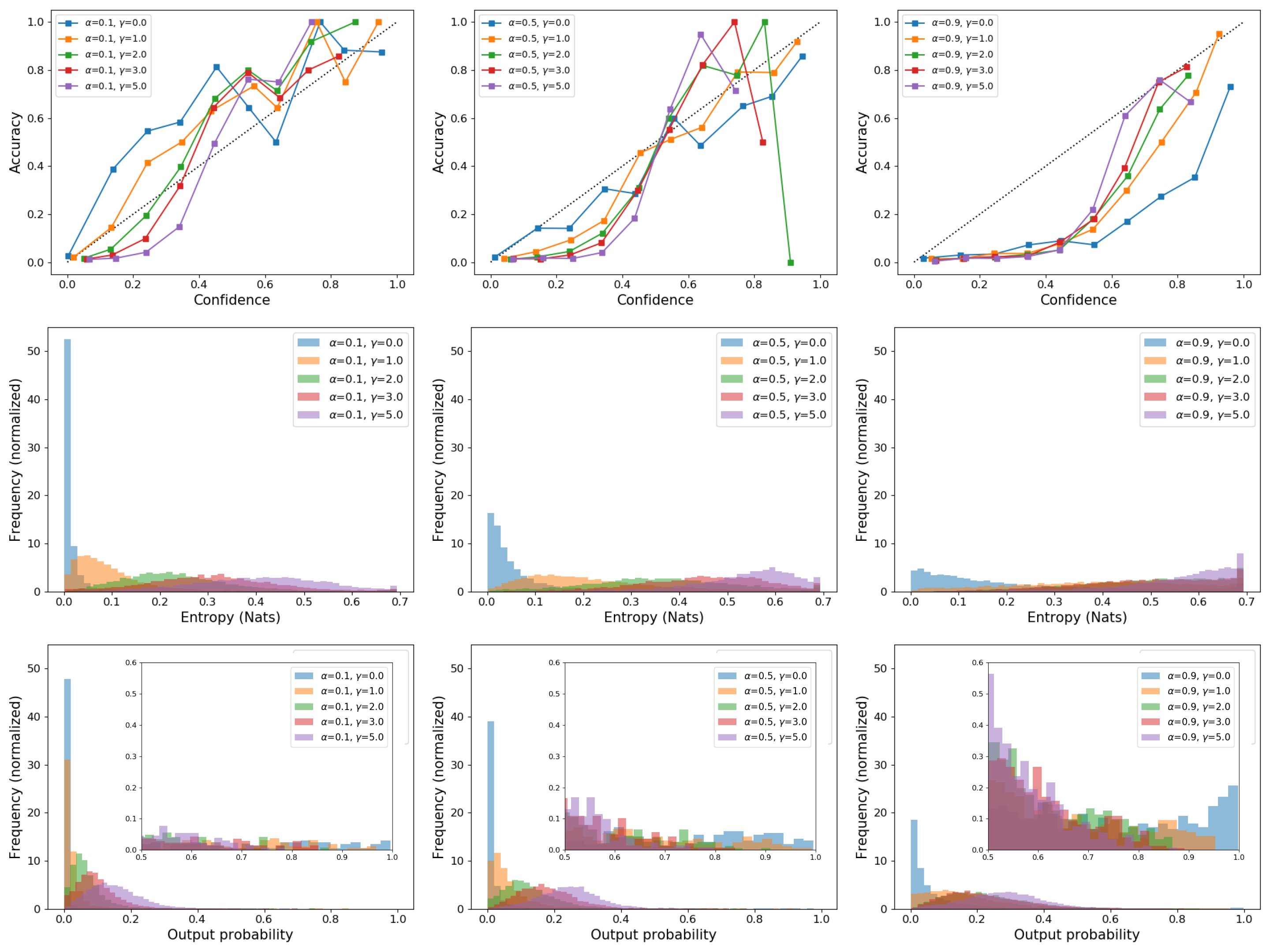}
\caption{The calibration curves (top), entropy histograms (middle), and output probability histogram (bottom) of different models trained with different hyperparameters in the focal loss.}
\label{fig:exp4_plots}
\end{figure}

As shown in eqn. \ref{eqn:FL}, weight factor larger than 0.5 gives larger penalty to misclassification of true positive samples, and weight factor smaller than 0.5 does the same to true negative samples.
As a result, using larger weight factor would encourage correct classification of true positive samples, yet misleading some negative samples to be classified as false positives. 
In other words, larger the weight factor, more the samples would be detected as positives; this would result in higher recall and lower precision values.

Such prediction is confirmed through our experiments. 
For models trained in different set of $\alpha_{\text{FL}}$ and $\gamma_{\text{FL}}$, Figure \ref{fig:exp4_summary} shows the prediction performance (accuracy, precision, recall, and F1-score) and the prediction reliability (ECE and OCE) results, and Figure \ref{fig:exp4_plots} shows corresponding calibration curves, entropy histograms, and output probability histograms. 
Varying the weight factor $\alpha_{\text{FL}}$ did significantly affect precision and recall. As expected, larger $\alpha_{\text{FL}}$ gave the model lower precision and higher recall in overall. 
In fact, F1-score, the harmonic mean of precision and recall, found best at $\alpha_{\text{FL}}=0.75$.

Now we assess and analyze the effect of focal loss on prediction performance and reliability. 
We could observe that focal loss did not improve the prediction performance, and even damaged the prediction reliability in all of our test cases except $\alpha=0.1$ and $\gamma=1.0$ case.   
We suspect that such detrimental effect of focal loss arises from the same reason ERL harms reliability; focal loss and ERL both push the predictive distribution to a uniform distribution by strongly penalizing high confidence predictions.
Since LS, ERL, and FL regularize predictive distribution itself, the predictive distribution cannot possibly estimate the true distribution without bias.

\subsection{Reliability of models in virtual screening scenario}

Lastly, we aim to imitate/cover a real-world virtual screening scenario - where screening library can be largely discrepant from training data distribution - by training models with DUD-E database\cite{mysinger2012directory} and testing the models on the ChEMBL database.\cite{gaulton2012chembl} Such experimental strategy is elaborated in the second experimental section of \citeauthor{ryu2019bayesian}\cite{ryu2019bayesian}.
Due to an inherent discrepancy between training and test data distribution, uncertainty of the prediction would be unavoidably higher in virtual screening situation. Thus, over-confident predictions would be exceptionally harmful, and predicting labels with correct probability estimation becomes significantly important to achieve high success rates.

We trained models by using the EGFR/VGFR2/ABL1 sets in the DUD-E database.
For each training, we built four different models -- baseline, MC-DO, LS, and ERL (same models in the second experiments). 
Then, we obtained the predictive probability of compounds associated to the EGFR/VGFR2/ABL1 sets in the ChEMBL database, where labels are given by negative log of half-maximum inhibitory concentration (pIC50) value. 
In order to set/view our problem of virtual screening as classification tasks, 
we let the label of compounds as negative (zero) if pIC50 is smaller than 7.0, and positive (one) otherwise.
In other words, we attempted to find the compounds whose pIC50 is larger than 7.0 with our model trained with the DUD-E dataset. 
More details on training procedure and datasets are provided in supplementary information. 

\begin{figure}[] 
    \includegraphics[width=0.99\textwidth,trim={0cm 0 0cm 0},clip]{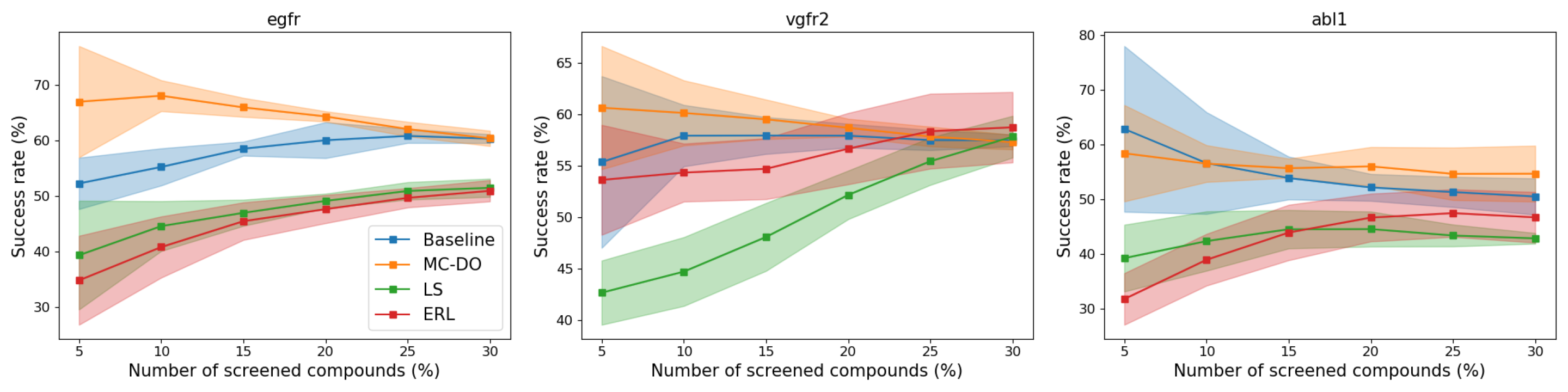}
    \caption{The change of success rate as varying the number of screened compounds for the scenarios of screening EGFR (left), VGFR2 (center), and ABL1 (right) active compounds.}   
    \label{fig:chembl_virtual_screening}
\end{figure}

We sorted the compounds by output probability in descending order, and screened the top K\%.   
Figure \ref{fig:chembl_virtual_screening} summarizes the success rate - the precision of prediction, or the ratio between a number of true positive compounds and a number of screened compounds - with respect to K value, for each model. 
If the models were well-calibrated, when we choose compounds of higher output probability, the success rate would be higher. For MC-DO model, the higher the output probability criteria became, the higher the success rate we achieved. However, the other models - LS and ERL - did not show such behaviour, providing considerably low success rate for screened top 5-10\% compounds. In that sense, we suggest MC-DO model as an appropriate model for virtual screening.

For more detailed analysis, we visualize the histogram of true positive, false positive, true negative, and false negative predictions of the four different models in Figure \ref{fig:egfr_tp_fp_tn_fn}, \ref{fig:vgfr2_tp_fp_tn_fn} and \ref{fig:abl1_tp_fp_tn_fn} in supplementary information. 
Since the regularizers of LS and ERL forces the predictive distribution similar to the uniform distribution, 
the models located a large amount of false positive compounds near 1.0.
We conjecture that such penalties lead to relatively low success rate for screening with probability criteria due to the biased probability estimation as described in the method section.

\subsection{Remarks} 

So far, we have presented a number of experimental results and analysis on prediction reliability of graph neural networks. 
We conclude our experimental analysis with the following remarks.

\textbf{Remark 1. ``Modest model capacity is necessary for reliable and accurate predictions."} 
Attention mechanism is now widely adopted for neural networks in various domains, and graph convolutional network (GCN) is not an exception. Accordingly, GAT - an attention adopted version of GCN - can easily be regarded as an advanced model for all-time purpose. 
However, we found out that GAT sometimes causes over-fitting and provides less reliable predictions, probably due to over-parameterization.
In the supporting information section, we provide the results of additional experiment where the models were trained with large number of training samples. 
Four types of model architectures used for the experiment were identical to those from aforementioned experiments, and those were trained for regression task. 
The results manifest that GAT model and attention readout did improve the prediction performance this time. 
The implication of such results is that depending on the size of dataset, proper model capacity should be chosen. 
Since data-deficient situations are very common in molecular applications, careful choice of model capacity must be even further emphasized.

\textbf{Remark 2. ``While regularization is necessary to improve prediction reliability, the careful choice of appropriate methods is essential."}
Our baseline model used L2-weight decay regularization, but it could be further improved by other regularizations. 
DO and MC-DO were effective in improving the prediction reliability thanks to the probabilistic nature of (approximate) Bayesian inference.
In that sense, the MC-DO model provided the highest success rate for screening compounds with probability thresholding. 
On the other hand, cost-sensitive learnings hurt the prediction reliability because it produces biased probability estimations. We observed that the LS and ERL models did not provide higher success rate of virtual screening.
Our demonstration highlights the clear importance of using appropriate regularization method in order to achieve reliable prediction and thus to attain the success in virtual screening. 

\textbf{Remark 3. ``Different weight factors in focal loss can bring changes in precision and recall."}
Since detecting rare samples - samples of a minority class - is much difficult than detecting abundant samples - samples of a majority class, the model often predicts a majority class sample with high output probability. 
Focal loss was initially proposed to handle such imbalanced data situation, because it penalizes easily predictable outcomes which typically show high output probabilities.
However, our study reveals that the improvement of either precision or recall, as well as F1-score, was mainly determined by the weight factor $\alpha_{\text{FL}}$.
Also, high $\gamma_{\text{FL}}$ resulted in poor prediction reliability due to the nature of cost-sensitive learning. 
Our findings highlight that weighted cross entropy, which is equivalent to the focal loss of $\alpha_{\text{FL}} \neq 0.5$ and $\gamma_{\text{FL}} > 0.0$, would be effective for handling imbalanced data. 

We believe that our findings give valuable lessons on developing virtual screening systems in different purposes. 
When ones desire to discover i) as many true positive samples as possible or ii) as less failures as possible, a model of i) high recall (few false negatives) or ii) high precision (few false positives) would be favorable for each scenario.
For example of toxicity prediction systems, models providing high recall performance can greatly reduce possible failures in clinical trials. 
In order to achieve either high precision or high recall model that allows reliable prediction,
our study recommends to: ``Do not penalize output probability, but use different weight factor.''

\section{Conclusion}

In this paper, we have presented the comprehensive study on the performance and reliability of graph neural networks in binary classification tasks of molecular properties. 
We followed the language of probability to describe the prediction reliability and assessed the reliability across models developed with various model architectures, regularizations, and loss functions. 
We concerned inevitable challenges in molecular applications, i.e. deficient and imbalanced data situation, and suggested a guide to achieve a model as reliable as possible -- ``Use modest model capacity, appropriate regularization and loss function, and learning and inference algorithm from Bayesian learning.'' 
Beyond our scope, we propose the following future research directions that expected to accomplish accurate and reliable prediction models.
\begin{itemize}
    \item There might be room for improvement in better model architectures for molecular graphs.
    For example, it would be valuable to study the usefulness of recent advancements in node pooling methods\cite{ying2018hierarchical, lee2019self} that reduce the dimensionality of node features.     
    While pooling is a common practice in convolutional neural networks for computer vision tasks, current graph neural networks based on message passing framework (such as GCNs) do not reduce the node feature dimensionality. Instead, graph neural networks simply aggregate all the node features, which sometimes result in producing graph features that lack node information.
    Hence, to enable better graph representation learning, we are keen to find an effective method to summarize statistics of node features: know-hows borrowed from convolutional neural networks (e.g. downsampling) might stand a chance.
    Eventually, it could leverage better predictions with less parameters. 

    \item More precise Bayesian learning algorithms would improve prediction reliability.
    Previous researches\cite{ryu2019bayesian, zhang2019bayesian} and this work have adopted MC-DO for approximate Bayesian inference due to the intractability in computing exact posterior distribution. 
    Since the uncertainty is estimated by the variance of predictive distribution, and predictive distribution is inferred by posterior distribution, 
    it is noteworthy to investigate the efficacy of advanced Bayesian learning methods in learning posterior distribution. 
    We believe that recent researches in Bayesian learning community\cite{mandt2017stochastic, mobiny2019dropconnect, osawa2019practical, maddox2019simple, wilson2020bayesian} can give fruitful hints for better Bayesian learning and reliable predictions. 
    
    \item Pre-trained models that enable better representation learning would also be beneficial for accurate and reliable predictions. \cite{devlin2018bert, he2019rethinking, hu2019pre, hendrycks2019using_pre, hendrycks2019using_self}
    Unsupervised representation learning has the virtue of label-free learning, and algorithms such as contrastive learning\cite{oord2018representation} facilitate obtaining representations useful for down-stream prediction tasks. \cite{henaff2019data}
    Since we can easily find abundant structural data of drug-like compounds from public chemical database\cite{irwin2005zinc, gaulton2012chembl}, such unsupervised pre-training methods can give an apt opportunity to develop models in data-efficient manners.
\end{itemize}
Consequently, we believe that our study will widen the opportunity of neural models in chemistry researches via reliable AI systems.


\section*{Acknowledgements}
We would like to appreciate Yongchan Kwon for his valuable comments on the effects of regularizations and experimental analysis.
This work was supported by the National Research Foundation of Korea (NRF) grant funded by the project 2019M3E5D4065965.

\section*{Author contributions}
S.R. and S. Y. conceived the idea and performed implementation and experiments. 
All the authors analyzed the results and wrote the manuscript together.

\section*{Conflicts of interest}
The authors declare no competing financial interests.

\bibliography{achemso-demo}

\pagebreak
\begin{suppinfo}

\subsection{A. Notes on choosing proper activation function for using the self-attention in graph nets}

\begin{figure}[] 
    \includegraphics[width=0.5\textwidth,trim={0cm 0 0cm 0},clip]{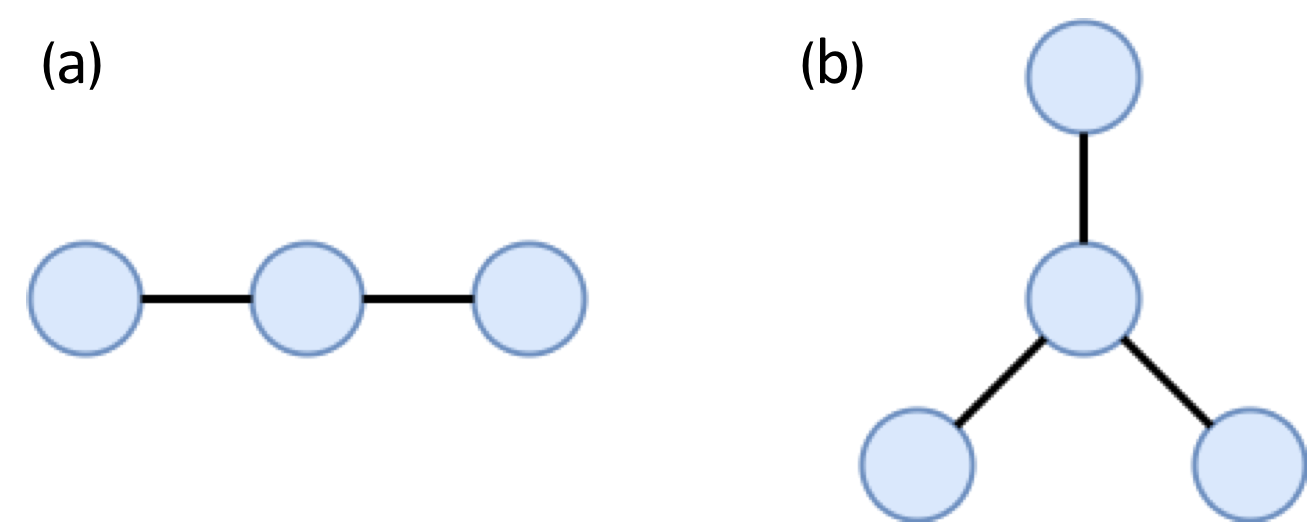}
    \caption{Two graphs (a) whose elements are three identical nodes and (b) four identical nodes.}
    \label{fig:invariance}
\end{figure}

Figure \ref{fig:invariance} visualizes two simple graphs consist of identical nodes but different numbers, which will be considered for our explanations on the importance of relevant update rules for node and graph featurizations. 

Firstly, we will consider using the self-attention in node featurizations, i.e. graph attention network, updating node features in the $(l+1)$-th node embedding layer by following the eqn. \ref{eqn:gat}. 
If we use softmax activation instead of tanh activation as described in eqn. \ref{eqn:gat_coeff}, 
node features will be updated as $(\frac{1}{3}h + \frac{1}{3}h + \frac{1}{3}h) = h$ and $(\frac{1}{4}h +  \frac{1}{4}h +  \frac{1}{4}h + \frac{1}{4}h) = h$ for each center node of two graphs shown in (a) and (b). 
This simple example tells us that using softmax activation which squashes the sum of logits to exactly one can lead to an identical node feature even their neighbor structures are different. 
This problem would not be problematic if node updating summarizes the distribution of neighbor nodes rather than the exact statistics.
Since node features in a molecular graph must reflect correct number and type of adjacent nodes, using softmax activation is notably a poor choice.
Thus, we used tanh activation and empirically found it shows better performance for all prediction tasks. 

Along with the same line, we can expand the above explanation for the graph featurization with attention readout. 
If we use softmax activation without scaling with the number of nodes $N$ as shown in eqn. \ref{eqn:attention_readout}, the attention readout aggregates node features to an identical graph feature, i.e. $z_{(a)} = (\frac{1}{3}h + \frac{1}{3}h + \frac{1}{3}h) = h$ and $z_{(b)} = (\frac{1}{4}h + \frac{1}{4}h + \frac{1}{4}h + \frac{1}{4}h ) = h$.
On the other hand, multiplying $N$ results to $z_{(a)} = 3h$ and $z_{(b)} = 4h$, which enables distinguish two different graph structures. 

\subsection{B. GAT show better prediction results than GCN when they are trained with large number of samples.}

\begin{table}[]
    \begin{tabular}{|c|c|c|c|c|}
    \hline
    \multicolumn{1}{|l|}{} & ZINC       \\ \hline
    Task type              & Regression \\ \hline
    Total training epoches & 50         \\ \hline
    Decay steps            & 20, 40     \\ \hline
    Number of samples      & 100,000    \\ \hline
    \end{tabular}
    \caption{Specifications on datasets and model training in the regression tasks}
    \label{tab:specification}
\end{table}

In the classification experiments, we observed that GAT (using the self-attention in GCN) damaged the both prediction performance and reliability. 
We further investigated whether the GAT with large amount of data samples can improve prediction ability or not.
We obtained octanol partition coefficient (logP), topological polar surface area (TPSA) and synthetic accessibility score (SAS) values by RDKit for each molecule in the ZINC dataset. 
We then trained regression models to predict the obtained values.
Hyper-parameters such as the total number of training epochs and steps to start learning rate decaying are noted in Table \ref{tab:specification}.
Each dataset was split to training set and test set by 80:20 ratio.

\begin{figure}[] 
    \includegraphics[width=0.75\textwidth,trim={0cm 0 0cm 0},clip]{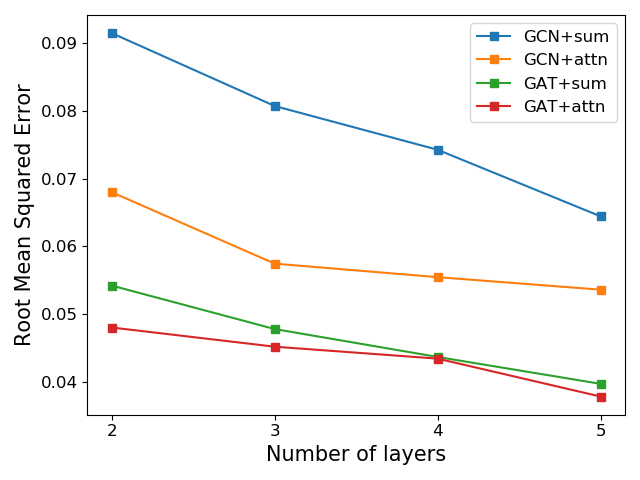}
    \caption{The change in logP prediction error as the number of graph convolution layers increases.}
    \label{fig:num_conv_layers}
\end{figure}

We trained logP prediction models with different node embedding (i.e. graph convolution and graph attention) and readout (i.e. sum and attention) methods. 
Figure \ref{fig:num_conv_layers} plots the change in logP prediction error in terms of root mean squared error (RMSE) as the number of node embedding layers increases.
The result confirms that applying attention mechanism for node embedding and readout outperforms other methods.

\begin{table}[]
    \begin{tabular}{|l|l|l|l|}
    \hline
                 & LogP           & TPSA          & SAS            \\ \hline
        GCN+sum  & 0.074          & 0.52          & 0.068          \\ \hline
        GCN+attn & 0.055          & \textbf{0.42} & 0.060          \\ \hline
        GAT+sum  & 0.044          & 0.53          & 0.057          \\ \hline
        GAT+attn & \textbf{0.043} & 0.52          & \textbf{0.053} \\ \hline
    \caption{The root mean squred error of logP, TPSA and SAS predictions.}
    \label{tab:zinc_predictions}
    \end{tabular}
\end{table}

Next, we evaluate the effect of model architecture on different prediction tasks. 
In this experiment, we used four node embedding layers.
Table \ref{tab:zinc_predictions} shows the RMSE of logP, TPSA, and SAS predictions.
Using attention mechanism for both node embedding and readout leads to the best performance except for TPSA prediction.

\subsection{C. Prediction results of the virtual screening experiments}

\begin{figure}[] 
    \includegraphics[width=0.9\textwidth,trim={0cm 0 0cm 0},clip]{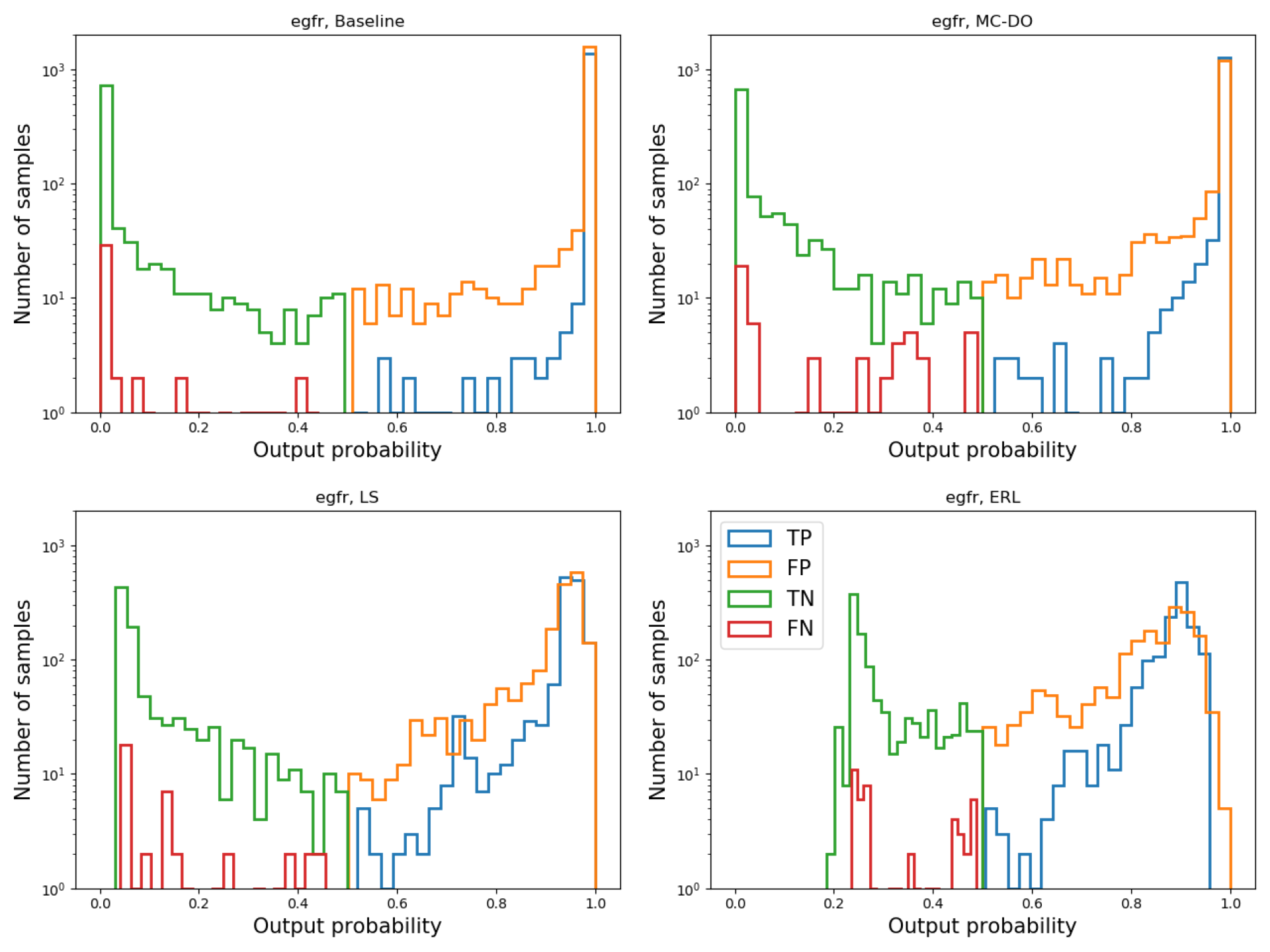}
    \caption{Distributions of output probability obtained by the baseline, MC-DO, LS, and ERL models for screening EGFR active compounds.
    The total distribution is divided into true positive (blue), false positive (orange), true negative (green) and false negative (red) groups. Note that the y-axis is represented with a log scale.}
    \label{fig:egfr_tp_fp_tn_fn}
\end{figure}

\begin{figure}[] 
    \includegraphics[width=0.9\textwidth,trim={0cm 0 0cm 0},clip]{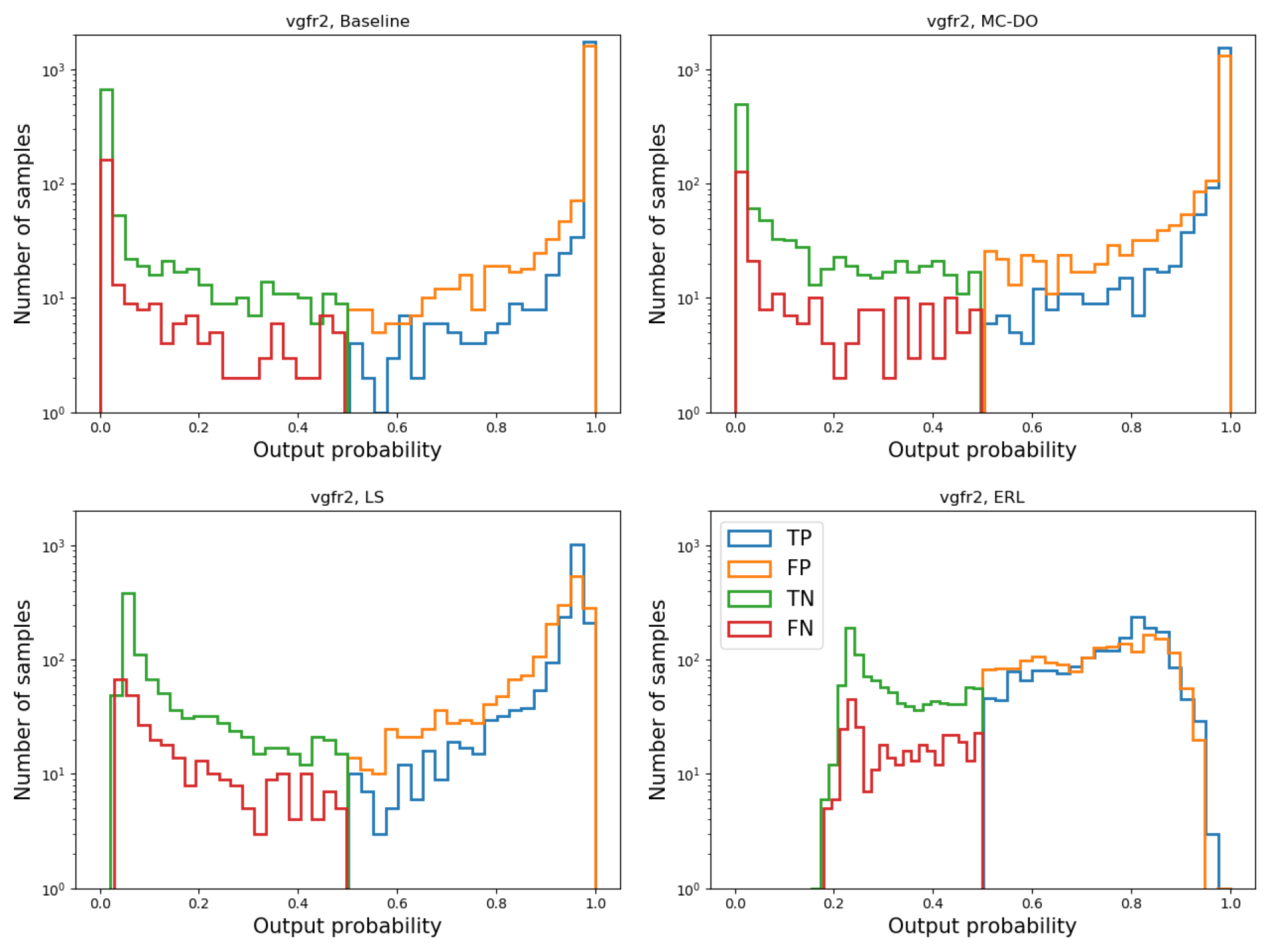}
    \caption{Distributions of output probability obtained by the baseline, MC-DO, LS, and ERL models for screening VGFR2 active compounds.
    The total distribution is divided into true positive (blue), false positive (orange), true negative (green) and false negative (red) groups. Note that the y-axis is represented with a log scale.}
    \label{fig:vgfr2_tp_fp_tn_fn}
\end{figure}

\begin{figure}[] 
    \includegraphics[width=0.9\textwidth,trim={0cm 0 0cm 0},clip]{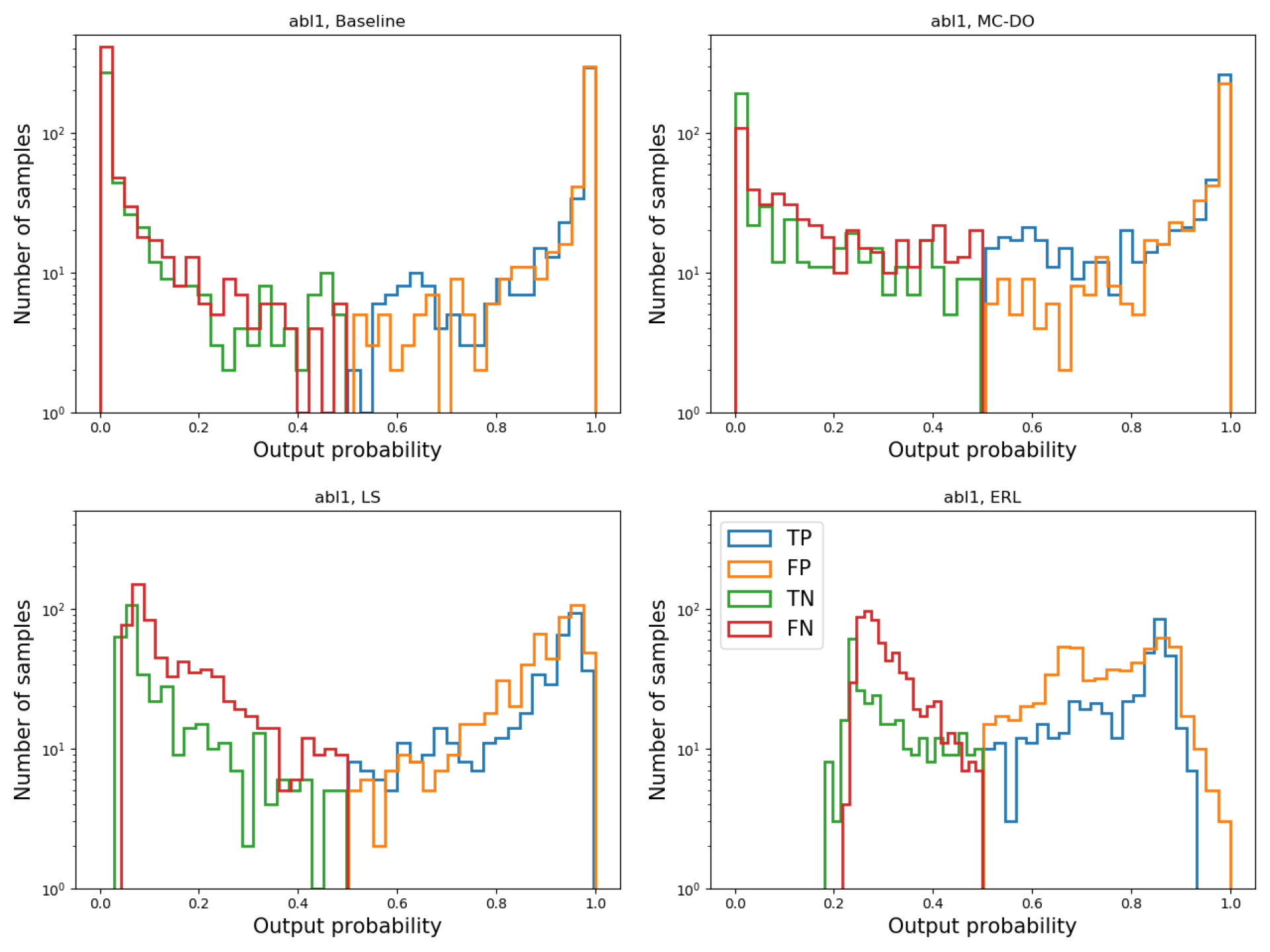}
    \caption{Distributions of output probability obtained by the baseline, MC-DO, LS, and ERL models for screening ABL1 active compounds.
    The total distribution is divided into true positive (blue), false positive (orange), true negative (green) and false negative (red) groups. Note that the y-axis is represented with a log scale.}
    \label{fig:abl1_tp_fp_tn_fn}
\end{figure}

\end{suppinfo}

\end{document}